# Towards a universal mechanism for successful deep learning

Yuval Meir[1,+], Yarden Tzach[1,+], Shiri Hodassman[1], Ofek Tevet[1] and Ido Kanter[1,2*]

[1]Department of Physics, Bar-Ilan University, Ramat-Gan, 52900, Israel.

[2]Gonda Interdisciplinary Brain Research Center, Bar-Ilan University, Ramat-Gan, 52900, Israel.

[+] These authors contributed equally

[*]Corresponding author email: ido.kanter@biu.ac.il

**Recently, the underlying mechanism for successful deep learning (DL) was presented based on a quantitative method that measures the quality of a single filter in each layer of a DL model, particularly VGG-16 trained on CIFAR-10. This method exemplifies that each filter identifies small clusters of possible output labels, with additional noise selected as labels outside the clusters. This feature is progressively sharpened with each layer, resulting in an enhanced signal-to-noise ratio (SNR), which leads to an increase in the accuracy of the DL network. In this study, this mechanism is verified for VGG-16 and EfficientNet-B0 trained on the CIFAR-100 and ImageNet datasets, and the main results are as follows. First, the accuracy and SNR progressively increase with the layers. Second, for a given deep architecture, the maximal error rate increases approximately linearly with the number of output labels. Third, similar trends were obtained for dataset labels in the range [3, 1,000], thus supporting the universality of this mechanism. Understanding the performance of a single filter and its dominating features paves the way to highly dilute the deep architecture without affecting its overall accuracy, and this can be achieved by applying the filter's cluster connections (AFCC).**

**Introduction**

A prototypical supervised learning task involves object classification, which is realized using deep architectures[1-3]. These architectures consist of up to hundreds of convolutional layers (CLs)[4-6], each of which consists of tens or hundreds of filters, and several additional fully connected (FC) hidden layers. As the classification task becomes more complex, a small training dataset and distant objects that belong to the same class, deeper architectures are typically required to achieve enhanced accuracies. The training of their enormous number of weights requires nonlocal training techniques such as backpropagation (BP)[7-9], which are implemented by advanced GPUs, and can guarantee convergence to a suboptimal solution only.

The current knowledge of the underlying mechanism of successful deep learning (DL) is vague[1,10-13]. The common assumption is that the first CL reveals a local feature of an input object, where large-scale features and features of features, which characterize a class of inputs, are progressively revealed in the subsequent CLs[1,14-17]. The terminologies of the features and features of features and the possible hierarchy among them have not been quantitatively well defined. In addition, the existence of the underlying mechanism of successful DL remains unclear. Is the realization of a classification task using deep and shallower architectures with different accuracies based on the same set of features? Similarly, is the realization of different classification tasks using a given deep architecture based on the same type of features?

A quantitative method to explain the underlying mechanism of successful DL[18] was recently presented and exemplified using a limited deep architecture and dataset, namely VGG-16[10] on CIFAR-10[14] and advanced variants thereof[10,19]. This method enables the quantification of the progressive accuracies with the layers and the functionality of each filter in a layer, and consists of the following three main stages.

In the first stage, the entire deep architecture is trained using optimized parameters to minimize the loss function. In the second stage, the weights of the first $m$ trained layers remain unchanged and their outputs are FC with random initial weights to the output layer, which represent the labels. The output of the first $m$ layers represents the preprocessing of an input using the partial deep architecture and the FC layer is trained to minimize the loss, which is a relatively simple computational task. The results indicate that the test accuracy[20] increases progressively with the number of layers towards the output.

In the third stage, the trained weights of the FC layer are used to quantify the functionality of each filter constituting its input layer. The single-filter performance is calculated with all weights of the FC layer silenced except for the specific weights that emerge from a single filter. At this point, the test inputs are presented and preprocessed by the first $m$ layers, but influence the output units only through the small aperture of one filter. The results demonstrate that each filter essentially identifies a small subset among the ten possible output labels, which is a feature that is progressively sharpened with the layers, thereby resulting in enhanced signal-to-noise ratios (SNRs) and accuracies[18]. These three stages, which constitute the method by which the performance of a single filter is calculated, are presented in Fig. 1.

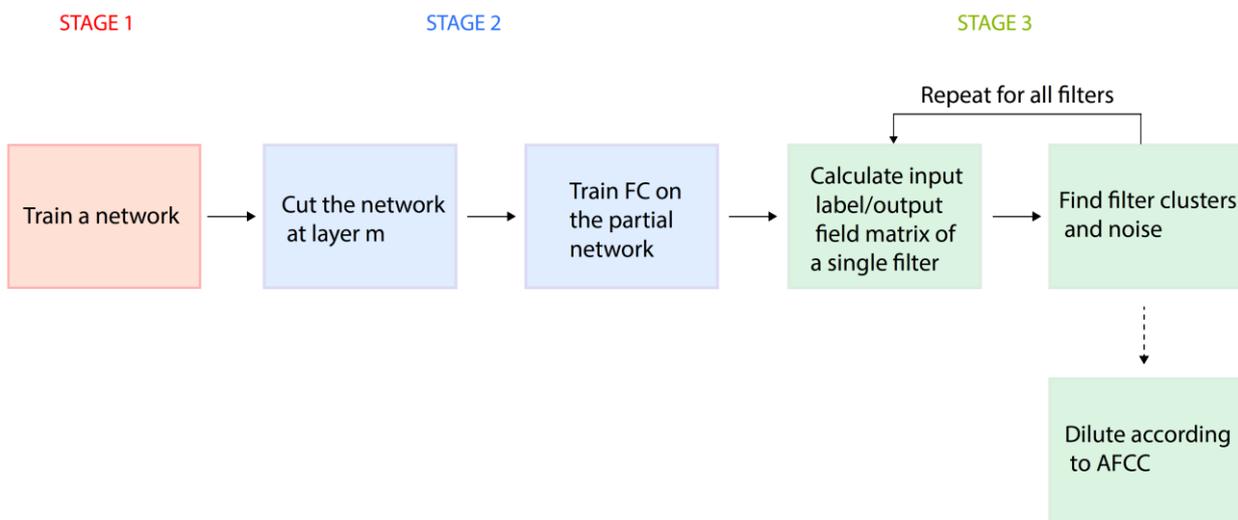

**Figure 1. Flowchart of the three stages for calculating the performance of a single filter.** The entire deep network is trained to minimize the loss function (Stage 1). The $m^{th}$ layer is FC to the output and is then trained to minimize the loss with fixed weights of the previous $m$ layers (Stage 2). The properties of a specific filter are calculated by silencing all the weights except those emerging from that specific filter. The matrix elements representing the average output field on an output unit for a specific input label are calculated using the training dataset. The clusters and noise elements of each filter are then calculated using the matrix elements. Finally, learning using a diluted deep architecture in accordance with the calculated clusters, namely the AFCC method, is performed.

As the method for the underlying mechanism of successful DL was tested for only one deep architecture and one dataset composed of small images[18], its generality is questionable. In this study, we investigate its universality by training EfficientNet-B0[21] and VGG-16 on extended datasets where the number of output labels is in the range of [3, 1,000], taken from CIFAR-10[14], CIFAR-100[14] and ImageNet[15,22]. The results strongly suggest the universality of the proposed DL mechanism, which is verified for varying numbers of output labels with three orders of magnitudes, small ($32 \times 32$) and large ($224 \times 224$) images, and state-of-the-art deep architectures.

In the following section, the underlying mechanism of DL is explained using the results for VGG-16 on CIFAR-100. Thereafter, the results are extended to EfficientNet-B0 on CIFAR-100 and ImageNet. Finally, the case of training VGG-16 and EfficientNet-B0 on varying number of labels taken from CIFAR-100 as well as VGG-16 on CIFAR-10 is discussed. Subsequently, a summary and several suggested techniques for improving the computational complexity and accuracy of deep architectures are briefly presented in the discussion section.

**Results**

**A. Results of VGG-16 on CIFAR-100**

The training of VGG-16 on CIFAR-100 (Fig. 2A) with optimized parameters yielded a test accuracy of approximately 0.75 (Table 1 and Supplementary Information), which was slightly higher than the previously obtained accuracy[23]. Next, the weights of the first $m$ trained layers were held unchanged, and their outputs were FC with random initial weights to the output layer. The selected layers were those that terminated with max-pooling, $m = 2, 4, 7, 10,$ and $13$. The training of these FC layers indicates that the accuracy increased progressively with the number of layers and saturated at $m = 10$, (Table 1), which is a result of the small image inputs of $32 \times 32$. The three CLs ($3 \times 3$), layers $8 - 10$, generate a $7 \times 7$ receptive field[24] covering a filter size of $4 \times 4$. Hence, layers $11 - 13$ are redundant for small images.

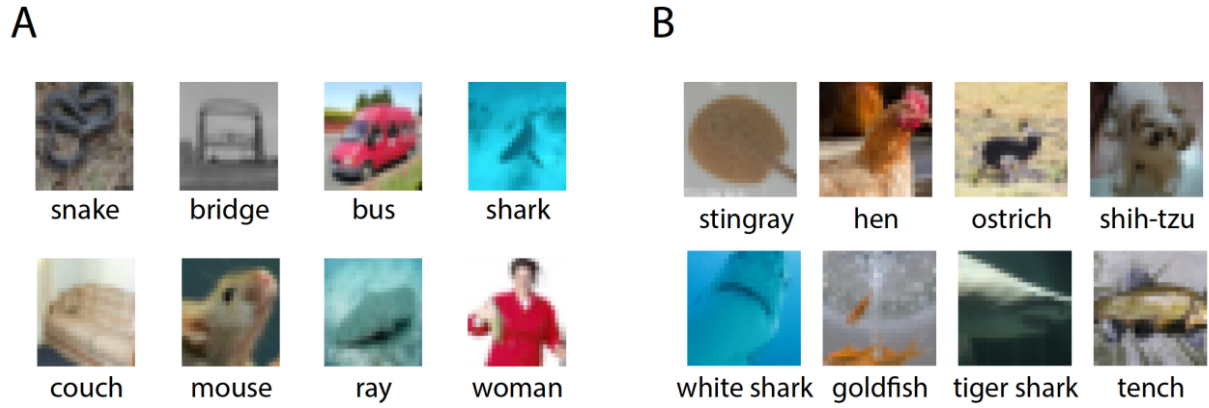

**Figure 2. Image samples of the datasets. (A)** Eight image samples with different labels from the CIFAR-100 dataset. **(B)** Eight image samples with different labels from the ImageNet dataset.

| VGG-16 on CIFAR-100 | | | | | | | |
|---|---|---|---|---|---|---|---|
| Layer | $N_f$ | $F_s$ | $FC_s$ | Accuracy | $n$ | $N_c$ | $C_s$ |
| 13 | 512 | 1x1 | 512 | 0.745 | 277.1 | 1.7 | 7.7 |
| 10 | 512 | 2x2 | 2048 | 0.752 | 16.3 | 2.6 | 2.0 |
| 7 | 256 | 4x4 | 4096 | 0.577 | 117.9 | 2.8 | 2.7 |
| 4 | 128 | 8x8 | 8192 | 0.439 | 552.4 | 5.1 | 2.8 |
| 2 | 64 | 16x16 | 16384 | 0.352 | 987.8 | 5.8 | 3.2 |

**Table 1: Accuracy per layer and statistical features of their filters for VGG-16 trained on CIFAR-100.** $N_f$: number of filters of layers terminating with max-pooling, $F_s$: filter sizes, $FC_s$: size of trained FC layer connected to the output units, $n$: average noise per filter, $N_c$: average number of clusters per filter, and $C_s$: average cluster size.

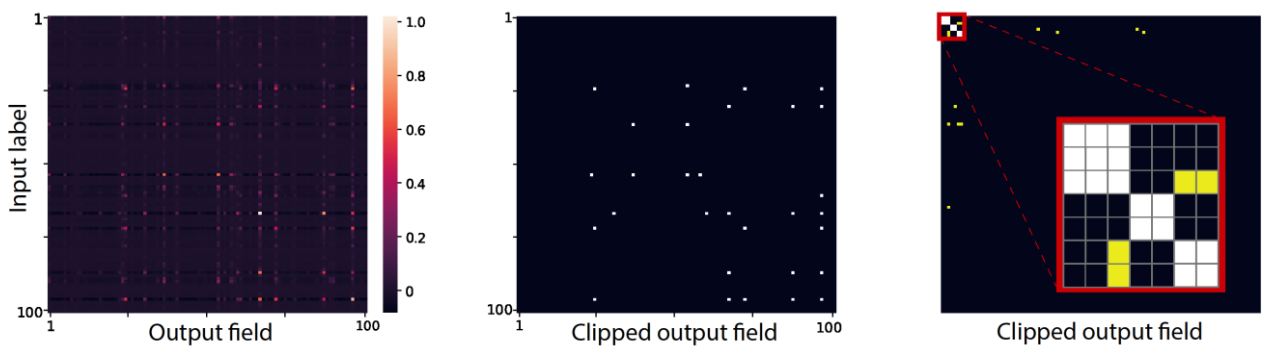

**Figure 3: Single filter performance.** Left: The matrix element $(i,j)$ of a filter belonging to layer 10 of VGG-16 trained on CIFAR-100 represents the averaged fields that were generated by label $i$ test inputs on an output $j$, where the matrix elements were normalized by their maximal element. Middle: The Boolean clipped matrix (0/1 is represented by black/white pixels) following a given threshold. Right: Permutations of the clipped matrix labels resulting in three diagonal clusters: two $2 \times 2$ and one $3 \times 3$ (magnified upper-left corner red box), where above-threshold $n$ elements out of the cluster are noise elements, denoted by yellow.

The performance of a single filter is represented by a $100 \times 100$ matrix and is exemplified for layer 10 (Fig. 3, left). The element $(i,j)$ represents the average of the fields that are generated by the label $i$ test inputs on output $j$, where the matrix elements are normalized by their maximal element. Next, its Boolean clipped matrix following a specified threshold is calculated (Fig. 3, middle) as well as its permuted version to form diagonal clusters (Fig. 3, right, Supplementary Information). The above-threshold elements out of the diagonal clusters are defined as the filter noise $n$ (yellow elements in Fig. 3, right).

The performance of each filter was calculated using test inputs, with all weights of the trained FC layer silenced except for those that emerged from the filter. The estimated main averaged properties of the $N_f(m)$ filters belonging to the $m^{th}$ layer are the cluster size $C_s(m)$, number of clusters per filter $N_c(m)$, and number of noise elements out of the clusters $n(m)$ (Table 1). The results clearly indicate that $n(m)$ decreases with $m$ until the accuracy is saturated at $m = 10$, where the average cluster size is small at 2 out of 100 labels. In addition, the average number of cluster elements is very small, $N_c \cdot C_s^2 = 2.6 \times 2^2 = 10.4$ out of the 10,000 matrix elements (Table 1).

The estimation of the SNR using the following quantities is required to understand the mechanism underlying DL. The average appearance number of each label among the $N_l$ labels in the clusters of the layer is

$$signal = (C_s \cdot N_c \cdot N_f)/N_l, \quad (1)$$

which represents the *signal* under the assumption of uniform number of appearances of each diagonal element over all clusters. The average expected *signal* that emerges from the 10th layer is approximately 26.6 (Table 1 and Eq. (1)), which fluctuates among the 100 labels (Fig. 4a). The average internal cluster noise, $noise_I$, is equal to the average number of appearances of other labels in the clusters forming the *signal* of a given label,

$$noise_I = \frac{(C_s - 1)}{N_l - 1} \cdot signal \quad (2)$$

which results in an average $noise_I$ of approximately 0.27 for the 10th layer, with relatively small fluctuations among the labels (Fig. 4a). Furthermore, $SNR_I = \frac{signal}{noise_I} \gg 1$ provided that $\frac{C_s}{N_l} \ll 1$.

The second type of noise stems from the above-threshold matrix elements out of the clusters, which is the external noise $n$. Using the assumption of uniform noise over the off-diagonal matrix elements, the average value of this noise is approximated as follows:

$$noise_E = n \cdot \frac{N_f}{(N_l)^2}, \quad (3)$$

where the average number of elements that belong to the clusters of each filter is negligible compared to $(N_l)^2$ (Fig. 3). As $noise_E \propto n$,

$$SNR_E = \frac{signal}{noise_E} = (C_s \cdot N_c \cdot N_l)/n, \quad (4)$$

which increases with a decrease in $n$. This is the origin of the DL mechanism, where $n$ decreases progressively with the number of layers, thereby enhancing the accuracy (Eq. (4)). For example, $noise_E$ is approximately 0.83 for the 10th layer, whereas it is approximately 29 for the 4th layer where the signal is only 18 (Table 1 and Eqs. (1)–(3)). Note that the above calculations neglect the subthreshold elements; however, they are typically several orders of magnitude smaller than the above-threshold elements and are frequently negative[18] (Fig. 3).

Although the above estimations of $SNR_I$ and $SNR_E$, Eqs. (1)–(4), were expected to fluctuate among the labels, they were found to be much greater than unity per label (Fig. 4a). In addition, these SNRs may be far from reality because the matrix (Fig. 3, left) was first normalized by its maximal value, which varied significantly among the filters, following which the above-threshold elements were defined to form a Boolean matrix. Nevertheless, the summation of the fields of the above-threshold elements, instead of their Boolean summations, indicates that $SNR_I$ and $SNR_E$ for each label were much greater than unity (Fig. 4b), and their averaged values are comparable to the estimated values based on the Boolean filters.

The progressive decrease in $noise_E$ with the layers of a given trained deep architecture is the underlying mechanism for successful DL (Eq. (4)). Nevertheless, a large estimated $SNR_E$ does not necessarily ensure an accuracy that approaches unity because it is based only on averaged quantities ((Eqs. (1)–(4)), where large fluctuations around their average values are expected, particularly for large $N_l$. In addition, a positive field of a cluster element cannot exclude negative fields for a large fraction of the corresponding input label.

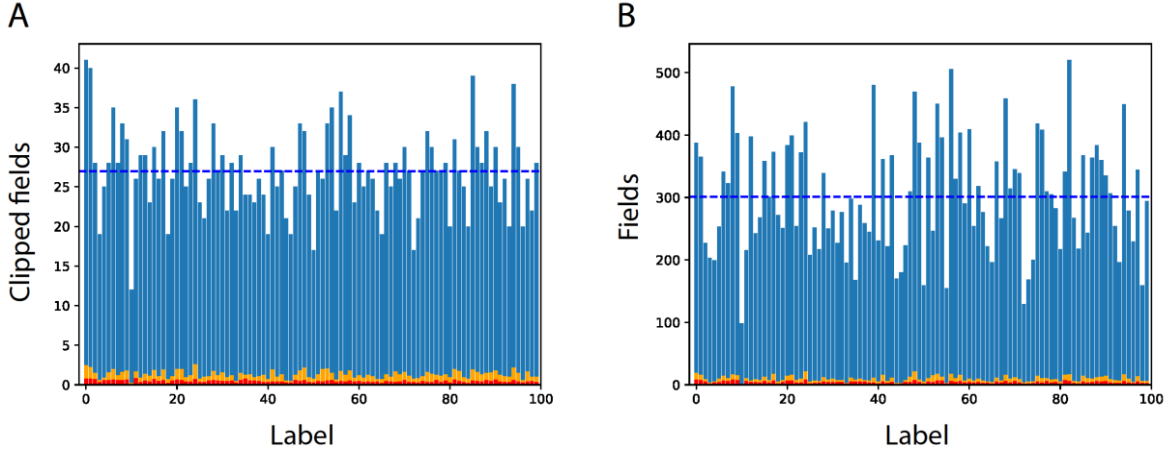

**Figure 4: Comparison of SNRs obtained from above-threshold Boolean filters and their fields. A.** The signal per label (blue), $noise_I$ per label (red), and $noise_I + noise_E$ per label (orange) (Eqs. (1)–(4)) that were obtained from the above-threshold clipped Boolean fields of the 512 filters of the 10$^{th}$ layer of VGG-16 trained on CIFAR-100. The average signal (dashed blue horizontal line), $noise_I$ (red), and $noise_I + noise_E$ (orange) are $26.95, 0.46$ and $1.3$, respectively, which are similar to the estimated values obtained from Eqs. (1)–(3). **B.** Similar to **A**, using the fields of the above-threshold elements of the filters. The average signal (dashed blue horizontal line), $noise_I$ (red), and $noise_I + noise_E$ (orange) values are $301, 4.7$, and $9.7$, respectively.

**B. Results of EfficientNet-B0 on CIFAR-100**

The training of the expanded $224 \times 224$ images[25] of CIFAR-100 on EfficientNet-B0 was performed using transfer learning[26,27] (Supplementary Information) and yielded an improved accuracy of $0.867$ (Table 2). This architecture does not include max-pooling operators, and a decrease of a factor of two in the layer dimensions is achieved using stride-2 at specific CLs. Hence, similar to the case of VGG-16, the accuracies and average filter properties were estimated at the end of the stages with stride-2, $1, 3, 4, 5, 7$, and $9$. The outputs of these stages were first sampled by $7 \times 7$ average pooling as built-in in stage 9, followed by a layer that was FC to the 100 output units which was trained to minimize the loss (Supplementary Information). The results indicate that the accuracy almost always increased with the number of stages and the noise per filter decreased (Table 2), thereby supporting the proposed universal mechanism underlying DL. The semi-plateau of the accuracies of stages 4 and 5 was common to all examined datasets using EfficientNet-B0, which suggests that this architecture might be simplified without affecting its accuracy by removing, for example, some layers around stage 5 (see Discussion section).

The progressive decrease in the noise $n$ with the layers or stages of a particular deep architecture is the underlying mechanism of DL. However, a comparison of the SNRs of two deep architectures does not necessarily correlate with accuracies. For instance, the improved EfficientNet-B0 accuracy of 0.867, in comparison with ~0.75 for VGG-16 (Tables 1–2), could not be simply deduced from their SNRs (Eq. (4)) because $noise_E$ was doubled for EfficientNet-B0, whereas $C_s \cdot N_c$ was reduced from 5.2 in VGG-16 to only approximately 4. The accuracy improvement of EfficientNet-B0 probably stems from the enhanced $signal$ of approximately 64, whereas it was only approximately 27 for VGG-16 (Eq. (1)), as well as the distribution of their output fields for the test inputs.

| EfficientNet-B0 on CIFAR-100 | | | | | | | |
|---|---|---|---|---|---|---|---|
| Stage | $N_f$ | $F_s$ | $FC_s$ | Accuracy | $n$ | $N_c$ | $C_s$ |
| 9 | 1280 | 1x1 | 1280 | 0.867 | 31.6 | 1.2 | 3 |
| 7 | 192 | 1x1 | 192 | 0.729 | 232.0 | 3.5 | 2.1 |
| 5 | 80 | 2x2 | 320 | 0.503 | 308.7 | 3.6 | 1.7 |
| 4 | 40 | 4x4 | 640 | 0.502 | 436.9 | 5.2 | 1.7 |
| 3 | 24 | 8x8 | 1536 | 0.426 | 526.2 | 5.6 | 1.8 |
| 1 | 32 | 16x16 | 8192 | 0.259 | 1208.1 | 2.9 | 5.3 |

**Table 2. Accuracy per stage and statistical features of their filters for EfficientNet-B0 trained on CIFAR-100.** The presented results were obtained at the end of stages consisting of stride-2 only, reducing by factor two the size of the output layer, similar to the max-pooling operator in VGG-16 (Table 1).

### C. Results of EfficientNet-B0 on ImageNet

The presented underlying mechanism of DL was extended to a dataset consisting of 1,000 labels and $224 \times 224$ input images, with the pre-trained EfficientNet-B0 on the

ImageNet dataset[15,22] (Fig. 2B) constituting the initial stage of the following procedure. The output layer of stages $1, 3, 4, 5, 7$, and $9$ was FC with random initial weights to the $1,000$ outputs (Table 3). Next, these FC weights were trained to minimize the loss, with all remaining weights of the trained EfficientNet-B0 kept fixed. Finally, the accuracy of the different stages and statistical properties of their filters were estimated (Table 3).

As training of these FC layers using the large ImageNet dataset ($1.4M$ images) was beyond our computational capability, we divided the $50,000$ images from the validation test into $40,000$ images for training and $10,000$ for testing. This training of the stage 9 FC layer was similar to transfer learning[26,27] and yielded an accuracy of approximately $0.75$, where the original accuracy of the entire pre-trained EfficientNet-B0 was approximately $0.78$ (Supplementary Information).

The accuracy increases with the stages, whereas the noise $n$ typically decreases (Table 3), which supports the universal underlying mechanism of DL. Interestingly, the average cluster size, $C_s$, and number of clusters per filter, $N_c$, which were measured at the last stage or layer that saturated the accuracy, increased only slightly while $N_l$ increased from 100 to $1,000$ (Tables 1–3). The exception of stage $3$ in which $n$ was non-monotonic (Table 3) may stem from the small $N_f = 24$, resulting in $N_f \cdot N_c \cdot C_s \sim 601 < 1,000$, whereas it was greater than $1,000$ for other stages. For stage 3, a large fraction of the labels ($\sim 500$) did not appear in any of the clusters and their estimated signal was zero. For all other stages, $N_f$ was larger and $N_f \cdot N_c \cdot C_s > 1,000$, resulting in significantly lower number of labels with zero signal. Note that this anomaly of stage 3 was indeed absent in CIFAR-100 (Table 1).

Similar trends are expected for VGG-16 on ImageNet with much lower accuracy and higher noise than EfficientNet-B0. In this case, the image dimension is greater by a factor of 7; hence, the FC layer sizes become significantly larger, and the optimization of those layers is currently beyond our computational capabilities.

| EfficientNet-B0 on ImageNet | | | | | | | |
|---|---|---|---|---|---|---|---|
| Stage | $N_f$ | Filter's outputs | FC size | Accuracy | $n$ | $N_c$ | $C_s$ |
| 9 | 1280 | 1x1 | 1280 | 0.750 | 1057.7 | 3.5 | 4.4 |
| 7 | 192 | 1x1 | 192 | 0.489 | 5729.6 | 8.3 | 3.2 |
| 5 | 80 | 2x2 | 320 | 0.187 | 7168.0 | 8.9 | 2.6 |
| 4 | 40 | 4x4 | 640 | 0.136 | 13416.9 | 16.5 | 2.5 |
| 3 | 24 | 8x8 | 1536 | 0.065 | 6471.0 | 13.2 | 1.9 |
| 1 | 32 | 16x16 | 8192 | 0.022 | 50381.9 | 8.5 | 7.4 |

**Table 3: Accuracy per stage and statistical features of their filters for EfficientNet-B0 trained on ImageNet**. The presented results were obtained at the end of stages consisting of stride-2 only, similar to Table 2.

## D. Datasets with varying number of labels

### D1. CIFAR-100 with varying number of labels

The proposed universal mechanism for DL was extended by varying the output labels $K$ out of $100$ in CIFAR-100, where $K = 10, 20, 40,$ and $60$. The results for VGG-16 are summarized in Table 4, and indicate similar trends to those observed for $K = 100$ (Table 1). The accuracy increased progressively with the number of layers until saturation at the $10^{th}$ layer, and the out-of-cluster noise $n$ decreased progressively with the number of layers. Interestingly, $C_s$ and $N_c$ were only slightly affected by $K$ at the 10th layer (Tables 1 and 4). The test error, $\epsilon = 1 - accuracy$, is expected to increase with $K$ since the classification task is more complex; the results indicate that this increase is approximately linear with $K$ (Fig. 5). Nevertheless, the extrapolation of the linear fit to a smaller $K$ approaching unity indicates that a limited crossover is expected, as $\epsilon$ is expected to vanish for $K = 1$.

| VGG-16 on CIFAR-10/100 | | | | | | | |
|---|---|---|---|---|---|---|---|
| Layer | $N_f$ | $F_s$ | $FC_s$ | Accuracy | $n$ | $N_c$ | $C_s$ |
| 13 | 512 | 1x1 | 512 | 0.926 | 3.26 | 1.01 | 2.2 |
| 10 | 512 | 2x2 | 2048 | 0.931 | 4.86 | 1.83 | 1.6 |
| 7 | 256 | 4x4 | 4096 | 0.908 | 10.11 | 1.47 | 1.7 |
| 4 | 128 | 8x8 | 8192 | 0.890 | 15.83 | 1.6 | 1.8 |
| 2 | 64 | 16x16 | 16384 | 0.829 | 18.64 | 1.6 | 2.0 |
| VGG-16 on CIFAR-20/100 | | | | | | | |
| Layer | $N_f$ | $F_s$ | $FC_s$ | Accuracy | $n$ | $N_c$ | $C_s$ |
| 13 | 512 | 1x1 | 512 | 0.9115 | 9.92 | 1.02 | 3.7 |
| 10 | 512 | 2x2 | 2048 | 0.9115 | 13.6 | 2.33 | 1.9 |
| 7 | 256 | 4x4 | 4096 | 0.9065 | 33.6 | 1.64 | 2.31 |
| 4 | 128 | 8x8 | 8192 | 0.8465 | 57 | 2 | 2.4 |
| 2 | 64 | 16x16 | 16384 | 0.752 | 68.23 | 1.83 | 2.7 |
| VGG-16 on CIFAR-40/100 | | | | | | | |

| Layer | $N_f$ | $F_s$ | $FC_s$ | Accuracy | $n$ | $N_c$ | $C_s$ |
|---|---|---|---|---|---|---|---|
| 13 | 512 | 1x1 | 512 | 0.8553 | 51.8 | 1.11 | 7.5 |
| 10 | 512 | 2x2 | 2048 | 0.8567 | 12.3 | 2.92 | 2 |
| 7 | 256 | 4x4 | 4096 | 0.7825 | 38.4 | 2.44 | 2.17 |
| 4 | 128 | 8x8 | 8192 | 0.6388 | 143.8 | 3.22 | 2.54 |
| 2 | 64 | 16x16 | 16384 | 0.5380 | 203.6 | 3.5 | 2.7 |
| **VGG-16 on CIFAR-60/100** | | | | | | | |
| Layer | $N_f$ | $F_s$ | $FC_s$ | Accuracy | $n$ | $N_c$ | $C_s$ |
| 13 | 512 | 1x1 | 512 | 0.8277 | 123.9 | 1.3 | 8.13 |
| 10 | 512 | 2x2 | 2048 | 0.8275 | 18.17 | 2.78 | 2.3 |
| 7 | 256 | 4x4 | 4096 | 0.7148 | 39.52 | 2.16 | 2.24 |
| 4 | 128 | 8x8 | 8192 | 0.5392 | 260.6 | 4.16 | 2.6 |
| 2 | 64 | 16x16 | 16384 | 0.4480 | 423.92 | 4.5 | 3 |

**Table 4: Accuracy per layer and statistical features of their filters for VGG-16 trained on $K$ labels from CIFAR-100.** The results are similar to those of Table 1, where VGG-16 was trained on $K = 10, 20, 30,$ and $60$ labels out of $100$, namely CIFAR-K/100 (Supplementary Information).

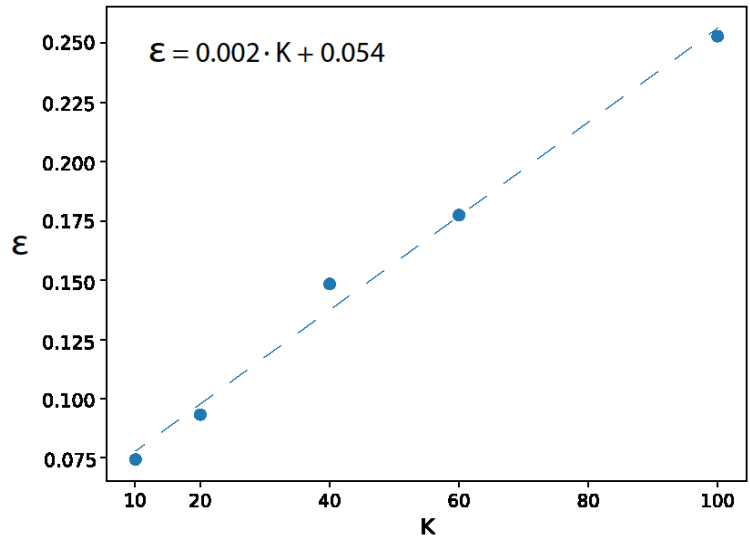

**Figure 5: Test error for VGG-16 trained on CIFAR-K/100.** Test error, $\epsilon = 1 - accuracy$, obtained at $10^{th}$ layer of VGG-16 trained on $K$ labels from CIFAR-100, namely CIFAR-K/100, and the linear fit approximation (dashed line). The subset of $K$ labels included smaller ($< K$) selected labels (Supplementary Information).

Similar trends were observed for EfficientNet-B0 trained with $K = 10, 20, 40,$ and $60$ labels from CIFAR-100 (Table 5). Again, the accuracy increased progressively with the stages (except for stage 5 at $K = 60$) and $n$ decreased progressively with the stages, thereby exemplifying the universality of the mechanism underlying DL. Similar to the case of VGG-16, the test error $\epsilon$ also increased approximately linearly with $K$ and almost vanished, as expected, at $K = 1$ (Fig. 6). Note that the slope of the approximated linear fit fluctuated slightly among the samples (Supplementary Information). In addition, the average cluster size $C_s$ increased slightly from 1.6 for $K = 10$ to 3 for $K = 100$, whereas the number of clusters per filter $N_c$ was approximately 1.1 and independent of $K$ (Table 5).

| EfficientNet-B0 on CIFAR-10/100 | | | | | | | |
|---|---|---|---|---|---|---|---|
| Stage | $N_f$ | $F_s$ | $FC_s$ | Accuracy | $n$ | $N_c$ | $C_s$ |
| 9 | 1280 | 1x1 | 1280 | 0.986 | 3.8 | 1.08 | 1.6 |
| 7 | 192 | 1x1 | 192 | 0.955 | 8.3 | 1.80 | 1.3 |
| 5 | 80 | 2x2 | 320 | 0.851 | 10.6 | 1.85 | 1.2 |
| 4 | 40 | 4x4 | 640 | 0.845 | 12.8 | 2.15 | 1.3 |
| 3 | 24 | 8x8 | 1536 | 0.755 | 14.5 | 2.75 | 1.3 |
| 1 | 32 | 16x16 | 8192 | 0.634 | 18.1 | 1.55 | 1.9 |
| EfficientNet-B0 on CIFAR-20/100 | | | | | | | |
| Stage | $N_f$ | $F_s$ | $FC_s$ | Accuracy | $n$ | $N_c$ | $C_s$ |
| 9 | 1280 | 1x1 | 1280 | 0.973 | 8.1 | 1.1 | 2.0 |
| 7 | 192 | 1x1 | 192 | 0.915 | 22.9 | 2.1 | 1.5 |
| 5 | 80 | 2x2 | 320 | 0.765 | 29.7 | 2.0 | 1.3 |
| 4 | 40 | 4x4 | 640 | 0.764 | 40.8 | 2.8 | 1.4 |
| 3 | 24 | 8x8 | 1536 | 0.645 | 48.1 | 3.4 | 1.4 |
| 1 | 32 | 16x16 | 8192 | 0.482 | 63.8 | 1.5 | 3.1 |
| EfficientNet-B0 on CIFAR-40/100 | | | | | | | |
| Stage | $N_f$ | $F_s$ | $FC_s$ | Accuracy | $n$ | $N_c$ | $C_s$ |
| 9 | 1280 | 1x1 | 1280 | 0.935 | 16.4 | 1.1 | 2.4 |
| 7 | 192 | 1x1 | 192 | 0.849 | 64.2 | 2.6 | 1.7 |
| 5 | 80 | 2x2 | 320 | 0.652 | 85.2 | 2.7 | 1.4 |
| 4 | 40 | 4x4 | 640 | 0.650 | 111.3 | 3.3 | 1.5 |
| 3 | 24 | 8x8 | 1536 | 0.553 | 129.6 | 3.9 | 1.6 |
| 1 | 32 | 16x16 | 8192 | 0.362 | 223.6 | 1.9 | 3.9 |
| EfficientNet-B0 on CIFAR-60/100 | | | | | | | |
| Stage | $N_f$ | $F_s$ | $FC_s$ | Accuracy | $n$ | $N_c$ | $C_s$ |
| 9 | 1280 | 1x1 | 1280 | 0.915 | 21.4 | 1.2 | 2.6 |
| 7 | 192 | 1x1 | 192 | 0.810 | 121.8 | 3.2 | 1.9 |
| 5 | 80 | 2x2 | 320 | 0.593 | 152.0 | 3.0 | 1.6 |
| 4 | 40 | 4x4 | 640 | 0.603 | 200.0 | 3.8 | 1.6 |
| 3 | 24 | 8x8 | 1536 | 0.511 | 252.3 | 4.8 | 1.7 |
| 1 | 32 | 16x16 | 8192 | 0.313 | 492.6 | 2.5 | 4.4 |

**Table 5: Accuracy per layer and statistical features of their filters for EfficientNet-B0 trained on $K$ labels from CIFAR-100.** The results here are similar to those of Table 2, where EfficientNet-B0 was trained on $K = 10, 20, 30,$ and $60$ labels out of $100$, namely CIFAR-K/100 (Supplementary Information).

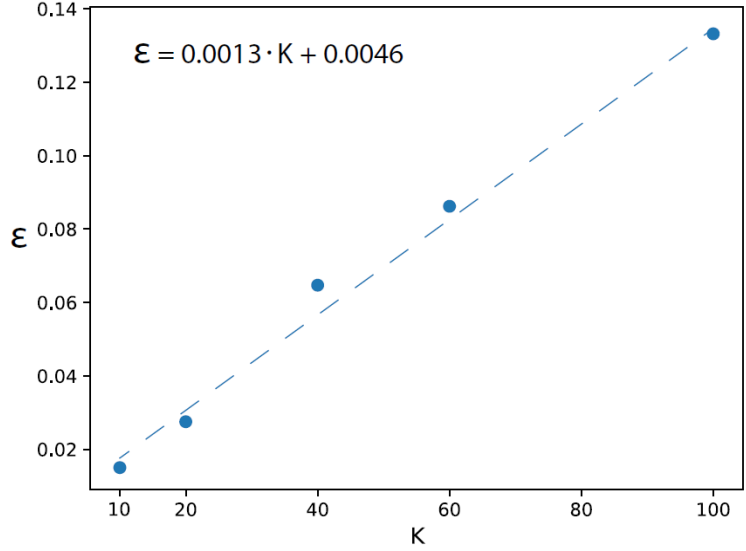

**Figure 6: Test error for EfficientNet-B0 trained on CIFAR-K/100.** Average test error, $\epsilon = 1 - accuracy$, obtained at stage $9$ of EfficientNet-B0 trained on $K$ labels from CIFAR-100 (similar to Fig. 5) and the linear fit approximation (dashed line).

### D2. CIFAR-10 with varying number of labels

The universal mechanism of DL was also verified for VGG-16 trained on CIFAR-10 with varying $K = 3, 6, 8,$ and $10$ (Table 6). The accuracy increased progressively with the number of layers until saturated at the 10$^{th}$ layer, and $n$ decreased progressively with the number of layers. Similar to the case of CIFAR-100, the test error increased approximately linearly with $K$ (Fig. 7), where the extrapolation for $K = 1$, $\epsilon$ approaches zero, as expected.

| VGG-16 on CIFAR-3/10 | | | | | | | |
|---|---|---|---|---|---|---|---|
| Layer | $N_f$ | $F_s$ | $FC_s$ | Accuracy | $n$ | $N_c$ | $C_s$ |
| 13 | 512 | 1x1 | 512 | 0.989 | 0.07 | 1 | 1.02 |
| 10 | 512 | 2x2 | 2048 | 0.989 | 0.27 | 1.5 | 1.02 |
| 7 | 256 | 4x4 | 4096 | 0.989 | 0.67 | 1.2 | 1.06 |
| 4 | 128 | 8x8 | 8192 | 0.972 | 1.70 | 1.1 | 1.12 |
| 2 | 64 | 16x16 | 16384 | 0.927 | 1.78 | 1.1 | 1.25 |
| VGG-16 on CIFAR-6/10 | | | | | | | |
| Layer | $N_f$ | $F_s$ | $FC_s$ | Accuracy | $n$ | $N_c$ | $C_s$ |
| 13 | 512 | 1x1 | 512 | 0.968 | 0.40 | 1 | 1.8 |
| 10 | 512 | 2x2 | 2048 | 0.967 | 1.16 | 2.4 | 1.3 |
| 7 | 256 | 4x4 | 4096 | 0.957 | 2.12 | 1.3 | 1.4 |
| 4 | 128 | 8x8 | 8192 | 0.930 | 6.69 | 1.2 | 1.6 |
| 2 | 64 | 16x16 | 16384 | 0.860 | 7.59 | 1.1 | 1.7 |
| **VGG-16 on CIFAR-8/10** | | | | | | | |
| Layer | $N_f$ | $F_s$ | $FC_s$ | Accuracy | $n$ | $N_c$ | $C_s$ |
| 13 | 512 | 1x1 | 512 | 0.961 | 0.63 | 1 | 2.2 |
| 10 | 512 | 2x2 | 2048 | 0.958 | 2.17 | 2.8 | 1.4 |
| 7 | 256 | 4x4 | 4096 | 0.954 | 4.07 | 1.2 | 1.6 |
| 4 | 128 | 8x8 | 8192 | 0.890 | 12.4 | 1.4 | 1.8 |
| 2 | 64 | 16x16 | 16384 | 0.783 | 13.0 | 1.3 | 1.8 |
| **VGG-16 on CIFAR-10/10** | | | | | | | |
| Layer | $N_f$ | $F_s$ | $FC_s$ | Accuracy | $n$ | $N_c$ | $C_s$ |
| 13 | 512 | 1x1 | 512 | 0.94 | 1.5 | 1 | 2.8 |
| 10 | 512 | 2x2 | 2048 | 0.94 | 3.8 | 3.2 | 1.6 |
| 7 | 256 | 4x4 | 4096 | 0.93 | 6.4 | 1.3 | 1.6 |
| 4 | 128 | 8x8 | 8192 | 0.85 | 18.3 | 1.4 | 2.1 |
| 2 | 64 | 16x16 | 16384 | 0.72 | 19.6 | 1.3 | 2.1 |

**Table 6: Accuracy per label and statistical features of their filters for VGG-16 trained on $K$ labels from CIFAR-10.** The results of VGG-16 trained on $K = 3, 6, 8,$ and $10$ labels, namely CIFAR-K/10 (Supplementary Information).

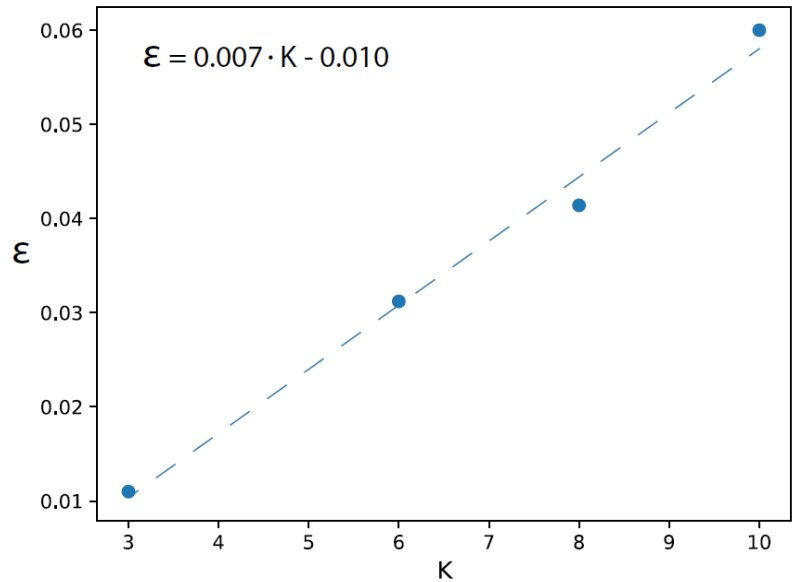

**Figure 7: Test error for VGG-16 trained on CIFAR-K/10.** Test error, $\epsilon = 1 - accuracy$, obtained at the 10th layer of VGG-16 trained on $K$ labels from CIFAR-10 and the linear fit approximation (dashed line).

**E. Applying Filter Cluster Connections (AFCC):**

The new comprehensive understanding of how the filters function in a trained deep architecture can promote improved technological implementation methods by applying filter's cluster connections (AFCC) (Fig. 1). As each filter consists of only several small clusters, thereby generating a significant output signal for a small set of labels, its output for any other label can be neglected and the same accuracy can be achieved. To test the AFCC hypothesis a trained VGG-16 on CIFAR-100 was examined, where the accuracy of approximately $0.752$, is saturated at the 10$^{th}$ layer (Table 1). The number of weights of the FC layer is $204,800$; $512 \times 2 \times 2$ input units emerging from the $512$ filters multiplied by $100$ output units. All these weights which did not belong to a cluster in a specific filter were set to zero, resulting in approximately $194,000$ zeroed weights out of $204,800$ (a $95\%$ reduction). The remaining $10,800$ weights is well approximated by $512 \cdot 2 \cdot 2 \cdot C_s \cdot N_c \approx 10,600$ (Table 1). After only a few training epochs, while maintaining the $\sim 194,000$ zeroed weights as zero, the similar accuracy, $\sim 0.752$, was recovered, which indicates that the FC layer can be significantly reduced and yield similar results (Supplementary Information). Note that the same filter clusters were detected for both training and test sets[18]. The performance of the same classification tasks with a significantly smaller amount of weights of the FC layer can improve the test computational complexity, as well as reduce the memory usage. Thus, the expansion of the AFCC method to include several layers can significantly reduce the complexity and deserves further research.

A similar effect was observed for EfficientNet-B0 trained on CIFAR-100, with an accuracy of $0.867$ (Table 2). The number of weights of the FC layer is $128,000$; $1,280 \times 1$ input units emerging from the $1,280$ filters multiplied by $100$ output units. All of these weights that did not belong to a cluster in a specific filter were set to zero, resulting in $4,900$ ($\sim 1280 \cdot C_s \cdot N_c$) non-zero weights only (a $\sim 96\%$ reduction). After retraining the entire network with the same parameters, while including the $4,900$ non-zeroed weights only, the accuracy increased to $\sim 0.873$, indicating that the FC layer can be significantly reduced and still yield similar or even increased accuracy (Supplementary Information). One cannot exclude a similar increase in accuracy without pruning the FC layer and using different training parameters, however, AFCC training is more efficient. This gain in the test computational complexity is expected to be enhanced further in datasets with a higher number of labels, such as ImageNet, and larger classification tasks.

The training of EfficientNet-B0 on CIFAR-100 indicates almost identical accuracies for stages 4 and 5 (Tables 2 and 5) whereas the noise, $n$, is non-monotonic between stages 3 and 4 for EfficientNet-B0 trained on ImageNet (Table 3). These results hint that stages $3-5$ of EfficientNet-B0 might be further optimized. Indeed, reducing the number of layers constituting stages 3 and 4 to one and training this modified EfficientNet-B0 on CIFAR-100 using transfer learning[26,27], resulted in an accuracy $\sim 0.864$, which approached the original accuracy (Table 2). Similarly, reducing the number of layers in stage 5 from 3 to 2, resulted in an accuracy of at least $0.862$ (Supplementary Information). Hence, following the proposed method, the latency of EfficientNet-B0 can be reduced without practically affecting its performance, at least for the CIFAR-100 dataset. Another simplification is the removal of stage 9 from the construction of EfficientNet-B0 and connecting stage 8 with only 320 filters to the output layer, using the AFCC method. In this case, the obtained accuracy is at least $0.868$, which slightly exceeds the accuracy of the entire model terminating with 1280 filters for the classification of CIFAR-100 (Supplementary Information).

## Discussion

The underlying mechanism of DL was quantitatively examined for two deep architectures, namely VGG-16 and EfficientNet-B0, trained on the CIFAR-10, CIFAR-100, and ImageNet datasets. These examinations enabled the verification of the suggested underlying mechanism of DL with different architectures consisting of 16 to over 150 layers as well as with the number of output labels ranging over three orders of magnitude [3,1,000].

The first step of the proposed method involves quantifying the accuracy of each CL of a trained deep architecture using the following procedure with relatively low computational complexity: The entire deep architecture is trained to minimize the loss. The weights of the first specified number of trained layers are held unchanged and their output units are FC to the output layer. These output units of an intermediate hidden layer represent the preprocessing of an input using a partial deep architecture, and the FC layer is trained to minimize the loss. The test set results indicate that the accuracy increases progressively with the number of layers towards the output (Tables 1–6).

The trained FC layer weights are used to quantify the functionality of each filter that belongs to its input layer. The single-filter performance is calculated when all weights of the FC layer are silenced, except for the specific weights that emerge from the single filter.

At this point, the test inputs are preprocessed by the first given number of trained layers, but influence the $N_l$ output units, representing the labels, only through the small aperture of one filter. This procedure generates an $(N_l, N_l)$ matrix, where element $(i, j)$ represents the average fields that are generated by label $i$ test inputs on output $j$. This matrix is normalized by its maximal element, following which a Boolean clipped matrix is formed following a given threshold. Its permuted version forms diagonal clusters (Fig. 3), the sizes of which increase only slightly when a deep architecture is trained on a dataset with an increasing number of labels (Tables 2 and 3). The diagonal elements of the clusters represent the signal, whereas their off-diagonal elements represent the internal noise, resulting in uncertainty regarding the input label given an above-threshold output. The second type of noise, namely the external noise, stems from the above-threshold elements out of the diagonal clusters. This noise progressively decreases with the number of layers and forms the underlying mechanism of DL.

The proposed method suggests quantitative measures and building blocks to describe the underlying mechanism of DL. The vocabulary is the preferred subset of labels of each filter clusters, which compete with the filter's noise. In addition to the contribution of this method to the understanding of how DL works, it provides insight into several practical aspects, including the following two. The first one is the possibility of improving the computational complexity and accuracy of deep architectures, and the second one is identifying weak stages in the construction of pre-existing deep architectures.

Using the single filter performance can lead to an efficient way to dilute the system without affecting its performance, as demonstrated by the AFCC method. Its expansion to include several layers can significantly reduce the complexity and deserves further research. This insightful dilution technique should be explored further on other datasets and deep architectures. In addition, its efficiency should be compared with that of other methods that primarily rely on random dilution processes[28-31] and assess their effectiveness in reducing complexity.

The presented universal underlying mechanism of DL may suggest an estimation method for the necessary number of filters in each layer. Each label must appear at least once in the clusters of the layer, hence, 1,280 filters in stage 9 of EfficientNet-B0 appear to be insufficient to classify, for example, 100,000 labels. Nevertheless, the results indicate that the number of diagonal elements, $C_s \cdot N_c$, increases from 3.6 for CIFAR-100 to 15.4 for ImageNet (Tables 2 and 3). Therefore, one cannot exclude the reality in which the filters constitute many relatively small clusters when the number of labels increases further. In addition, the information that is embedded in a single filter, namely clusters and noise, suggests procedures for pruning or retraining inefficient filters, such as highly noisy or low output-field filters. These procedures may improve the accuracy with reduced computational complexity and latency in the test phase, however, the investigation thereof requires further research.

# Supplementary Information

# Towards a universal mechanism for successful deep learning

Yuval Meir[1,+], Yarden Tzach[1,+], Shiri Hodassman[1], Ofek Tevet[1] and Ido Kanter[1,2*]

[1]Department of Physics, Bar-Ilan University, Ramat-Gan, 52900, Israel.

[2]Gonda Interdisciplinary Brain Research Center, Bar-Ilan University, Ramat-Gan, 52900, Israel.

[+] These authors contributed equally

[*]Corresponding author email: ido.kanter@biu.ac.il

**Architectures and Training the fully connected layer.** Two different architectures were examined. VGG-16[1] and EfficientNet-B0[2]. Both architectures were trained to classify the CIFAR-10 and CIFAR-100 datasets, as well as subclasses of their labels. In addition, EfficientNet-B0 was trained to classify the ImageNet dataset. Both architectures were trained with no biases on the output units. This was done to assure that each filter's effect on the output fields will be exemplified and will not be overshadowed by the much larger biases. Removing the biases of the output layer did not affect the architectures' average accuracies, in comparison to architectures trained with output biases.

The examination process was done by taking each architecture at designated layers and training a fully connected (FC) layer between the output of that specific layer and the output layer, corresponding to the labels. During training, only the FC layer was trained, while weights and biases of the rest of the architecture remain fixed. For VGG-16 the input units to the FC layers were selected after the max-pooling operations adjacent to layer 2, 4, 7, 10, and 13. For EfficientNet-B0, stages 1, 3, 4, 5, 7 and 9 which reduce the input size due to the stride-2 were examined.

For each examined layer, $m$, the output of the training set for the $m^{th}$ layer was used as a preprocessed dataset to train the FC layer. For each architecture, optimized hyper-parameters were used for the examined layers.

**Data preprocessing.** For VGG-16, each input pixel of an image $(32 \times 32)$ from the CIFAR-10 and CIFAR-100 databases was divided by the maximal pixel value, 255, multiplied by 2, and subtracted by 1, such that its range was $[-1, 1]$. In all simulations, data augmentation was used, derived from the original images, by random horizontally flipping and translating up to four pixels in each direction.

For EfficientNet-B0, the images were normalized by subtracting the average value of each color and dividing by its standard deviation. This varies by the size of the training set that was used, which change based on the number of different labels trained, $K$. For CIFAR-K/10 and CIFAR-K/100 the images were also expanded from their initial size of $(32 \times 32)$ to $(224 \times 224)$[3]. For all datasets, data augmentation was also used, which included a random horizontal flip, a random rotation of up to two degrees, a random translation of the image of up to four pixels in each direction and a shear of up to two degrees.

**Optimization.** The cross-entropy cost function was selected for the classification task and was minimized using the stochastic gradient descent algorithm[4,5]. The maximal accuracy was determined by searching through the hyper-parameters (see below). Cross-validation was confirmed using several validation databases, each consisting a fifth of the training set examples, randomly selected. The averaged results were in the same standard deviation (Std) as the reported average success rates. The Nesterov momentum[3] and L2 regularization method[4] were applied.

**Hyper-parameters.** The hyper-parameters η (learning rate), μ (momentum constant[6]), and α (regularization L2[4]) were optimized for offline learning, using a mini-batch size of 100 inputs. The learning rate decay schedule[5,7] was also optimized. A linear scheduler was used such that it was multiplied by the decay factor, q, every $\Delta t$ epochs, and is denoted below as $(q, \Delta t)$. Different hyper-parameters were used for each one of the architectures on each classification task.

**Datasets.** The used datasets were CIFAR-10, CIFAR-100 and ImageNet. Tests on the systems were extended by creating smaller datasets of $K$ labels, chosen form the CIFAR-10 and CIFAR-100 datasets. For CIFAR-100, $K = 20, 40, 60,$ and $100$ were used where for each progressively increasing K the previous subset is included, e.g. the labels chosen for $K = 20$ are included in $K = 40$, and for CIFAR-10, $K = 3, 6, 8,$ and $10$ were used. The CIFAR-K/10 and CIFAR-K/100 were normalized like CIFAR-10 and CIFAR-100 respectively for each architecture.

For ImageNet, 10,000 images, 10 images per label, were selected out of the validation set as the test dataset and the remaining 40,000 as training images.

**VGG-16 Hyper-parameters.**

VGG-16 was trained using the following hyper-parameters to reach maximal accuracies on CIFAR-K/100:

| VGG-16 | | | |
| --- | --- | --- | --- |
| CIFAR-20/100 | | | |
| η | μ | α | epochs |
| 0.004 | 0.965 | 3e-3 | 300 |

| CIFAR-40/100 | | | |
|---|---|---|---|
| η | μ | α | epochs |
| 0.002 | 0.975 | 4e-3 | 300 |
| CIFAR-60/100 | | | |
| η | μ | α | epochs |
| 0.002 | 0.975 | 4e-3 | 300 |
| CIFAR-100 | | | |
| η | μ | α | epochs |
| 0.002 | 0.975 | 4e-3 | 300 |

Where the decay schedule for the learning rate is:

$$(q, \Delta t) = (0.65, 20)$$

**For the training of the FC layer.** Each layer $m$ was FC to the $K$ outputs via a FC layer of size $N(m) \cdot K$. The FC layer was trained using the hyper-parameters: $\eta = 0.005$, $\mu = 0.975$, $\alpha = 1.5e - 5$, with a learning rate scheduler of $q = 0.65$ every 20 epochs while the rest of the system's weight values and biases remained fixed.

VGG-16 was trained using the following hyper-parameters to reach maximal accuracies on CIFAR-K/10:

| VGG-16 | | | |
|---|---|---|---|
| CIFAR-K/10 | | | |
| η | μ | α | epochs |
| 0.01 | 0.975 | 0.0015 | 200 |

Where the decay schedule for the learning rate is:

$$(q, \Delta t) = (0.65, 20)$$

**For the training of the FC layer,** $\eta = 0.02$, $\mu = 0.995$, $\alpha = 1e - 7$, with a learning rate scheduler of $q = 0.6$ every 20 epochs while the rest of the architecture's weight values and biases remained fixed.

**EfficientNet-B0 Hyper-parameters.**

EfficientNet-B0 was trained on CIFAR-K/100 and ImageNet datasets using transfer learning[8] on the pre-trained EfficientNet-B0 on ImageNet dataset. The transfer learning was done using the following hyper-parameters and learning rate scheduler:

| EfficientNet-B0 | | | |
|---|---|---|---|
| CIFAR-K/100 | | | |
| $\eta$ | $\mu$ | $\alpha$ | epochs |
| 0.01 | 0.9 | 0.001 | 200 |

Where the decay schedule for the learning rate is:

$$(q, \Delta t) = (0.975, 1)$$

For the first seven stages, the learning rate $\eta$ was multiplied by a factor of 0.1, and for the last stage by 0.2.

The output of each layer $m$ was sampled by a $7 \times 7$ average-pooling and then FC to the $K$ outputs via a FC layer of size $N(m) \cdot K$. The FC layer was trained using the hyper-parameters: $\eta = 0.005$, $\mu = 0.975$, $\alpha = 1.5e - 5$, with a learning rate scheduler of $q = 0.975$ every epoch while the rest of the architecture's weight values and biases remained fixed.

The training of EfficinetNet-B0 with reduced number of layers in stages 3 and 4 to 1, was done by using the hyper-parameters: $\eta = 0.002$, $\mu = 0.98$, $\alpha = 1e - 4$.

The training of EfficinetNet-B0 with reduced number of layers in stage 5, from 3 to 2, was done by using the hyper-parameters: $\eta = 0.01$, $\mu = 0.965$, $\alpha = 5e - 4$.

**Explanations for Figure 1.** In Fig 1 Left column, For VGG-16 on CIFAR-100, the 100 output fields of each filter were summed over all 10,000 inputs of the test set, resulting in a $100x100$ matrix where each cell $(i, j)$ represents the summed field of output field $j$ for all test set inputs of label $i$. The matrix was then normalized by dividing by its maximal value, resulting in each matrix having a maximal value of 1. In the center column the clipped Boolean output field matrix is displayed, where each element whose value is above a threshold (0.3) is set to 1 and all others are zeroed.

In the right column, the axes are permuted such as all labels belonging to a cluster are grouped together consecutively, thereby displaying the clusters in an adjacent fashion where they are displayed as a diagonal block of elements with value 1. Each cluster is defined as a subset of $n$ indices where for each $i, j \in n$ elements $(i, j)$ have the value of 1. The size of the cluster is defined as $n^2$ where $n$ is the number of labels whose all possible pair permutations form the cluster, where the minimal size can be 1, that is one element on the diagonal or 100, the entire matrix. The elements that are equal to 1 are then colored as white, representing that they belong to a cluster in the filter, while non-cluster cells with the value of 1 are classified as above-threshold external noise and are colored yellow.

The calculation of the clusters was done by running along the diagonal, from index (0,0) to (99,99) where the first $(i, i)$ element to have a value of 1 is initially designated as a cluster of size $1x1$. The next $(j, j)\ where\ j \neq i$ element to have a value of 1 is then checked to see if can complete a cluster with $(i, i)$, if yes, then it is added to the cluster and the next diagonal element to have a value of 1 is checked if completes a cluster with $i$ and $j$, if yes it is appended to the cluster, if not the system continues to the next cell. This process is repeated for all value 1 cells in the diagonal as long as there are elements who do not belong to a cluster. Note that this process is not uniquely defined, the order by which the indices are iterated can change the outcome of the clustering process, such as a filter with two clusters of sizes $3x3$ and $1x1$ retrieved by iterating from 0 to 99 can yield in certain very rare scenarios, two clusters of size $2x2$. While possibly alternating the results of a single filter, the overall obtained averaged results remain the same when performing the cluster creation while iterating in a reversed order, since those scenarios are very rare and occur in a negligible number of filters.

The external noise is calculated for each filter as the elements with value 1 who do not belong to any cluster. They can be seen in color yellow in the right column.

**Explanations for Figure 2.** **A.** The clipped binary signal per label was obtained by the diagonal signal of the summation of all binary clipped matrices of the filters together. The $noise_I$, the average internal noise of each label, is equal to the sum of all non-diagonal elements belonging to a cluster, on that label's row. The external noise, $noise_E$, of each label is equal to the sum of all non-diagonal elements not belonging to a cluster, on that label's row. **B.** Similar to **A** but now the signal, internal noise and external noise were calculated by the original accumulated fields of the filters and not the clipped binary fields.

The internal and external noise were summed by their obtained unit indices from the binary clipped matrix.

**Figure 3. Test error for VGG-16 trained on CIFAR-K/100** The error rate of VGG-16 on CIFAR-K/100 was tested with $K = 20, 40, 60$ and $100$, where the subset for the lowest value of K labels were randomly chosen and then for progressively increasing K the previous K labels were included, e.g. the labels chosen for $K = 20$ are included in $K = 40$. The K labels were chosen uniformly from the 20 super-classes of the dataset.

**Figure 4. Test error for EfficientNet-B0 trained on CIFAR-K/100** The error rate of EfficientNet-B0 on CIFAR-K/100 was tested with $K = 20, 40, 60$ and $100$, as done in Figure 3. The slope of the fitted line for the accuracies was 0.0013. This process was repeated for 5 different subsets and the slopes fluctuated in the range [0.0012, 0.0013].

**Figure 5. Test error for VGG-16 trained on CIFAR-K/10** The error rate of VGG-16 on CIFAR-K/10 was tested with $K = 3, 6, 8$ and $10$, where the subset for the lowest value of K labels were randomly chosen and then for progressively increasing K the previous K labels were included, e.g. the labels chosen for $K = 3$ are included in $K = 6$.

**Applied Filter's Cluster Connections (AFCC).** Optimization of the system can be achieved by capitalizing on the nature of the filter clusters. First, each output unit only receives a signal from filters whose clusters contain that output label, meaning that diluting all other unnecessary connections can reduce the complexity of the system. Second, filters can be trained by inputs whom belong to their clusters, thereby lowering the complexity of the training as well as the number of computational tasks needed. The first point is exemplified in the Discussion, where a trained FC was connected to the 10th layer, yielding an accuracy of ~0.752 and all weights who connect an output unit with a filter that does not constitute its filter are set to zero. After zeroing all FC weights, a short training section of a few epoch ~10 is commenced, where the zeroed weights remain zero and only the FC and 10th convolution layer are trained. The hyper-parameters used were $\eta = 0.00001$, $\mu = 0.9$, $\alpha = 1.5e - 2$. A similar effect was observed for EfficientNet-B0 on CIFAR-100 where each output unit only receives a signal from filters whose clusters contain that output label, but also the entire network was further trained, yielding an accuracy of ~0.873. The hyper-parameters used were $\eta = 0.005$, $\mu = 0.975$, $\alpha = 1e - 4$. For the first seven stages, the learning rate $\eta$ was multiplied by a factor of 0.1, and for the last stage by a factor of 0.2. The decay schedule for the learning rate was $(q, \Delta t) = (0.975, 1)$.

**Statistics.** Statistics for all results of EfficientNet-B0 were obtained using at least five samples. For VGG-16, results on CIFAR-K/10 were obtained using five samples, while for CIFAR-K/100 results were obtained using at least four samples.

**Hardware and software**. We used Google Colab Pro and its available GPUs. We used Pytorch for all the programming processes.